\documentclass[sigconf]{acmart}
\usepackage{bm,flushend,multirow}
\usepackage{appendix}
\usepackage{etoolbox}

\AtBeginDocument{%
  \providecommand\BibTeX{{%
    \normalfont B\kern-0.5em{\scshape i\kern-0.25em b}\kern-0.8em\TeX}}}

\copyrightyear{2021}
\acmYear{2021}
\setcopyright{acmlicensed}
\acmConference[MileTS '21]{MileTS '21: 6th KDD Workshop on Mining and Learning from Time Series}{August 14th, 2021}{Singapore}
\acmBooktitle{MileTS '21: 6th KDD Workshop on Mining and Learning from Time Series, August 14th, 2021, Singapore}
\acmPrice{15.00}
\acmISBN{978-1-4503-9999-9/18/06}

\begin{document}

\title[Submission and Formatting Instructions for MileTS'20]{Improving COVID-19 Forecasting using eXogenous Variables}

\author{Mohammadhossein Toutiaee}
% \authornotemark[1]
\email{hossein@uga.edu}
\affiliation{%
  \institution{Department of Computer Science \\
  The University of Georgia}
  \city{Athens}
  \state{GA}
}

\author{Xiaochuan Li}
% \authornotemark[1]
\email{xiaochuan.li@uga.edu}
\affiliation{%
  \institution{Department of Statistics \\
  The University of Georgia}
  \city{Athens}
  \state{GA}
}

\author{Yogesh Chaudhari}
% \authornotemark[1]
\email{yogesh.chaudhari@uga.edu}
\affiliation{%
  \institution{Department of Computer Science \\
  The University of Georgia}
  \city{Athens}
  \state{GA}
}

\author{Shophine Sivaraja}
% \authornotemark[1]
\email{shophine@uga.edu}
\affiliation{%
  \institution{Department of Computer Science \\
  The University of Georgia}
  \city{Athens}
  \state{GA}
}

\author{Aishwarya Venkataraj}
% \authornotemark[1]
\email{aishwaryavenkatraj@uga.edu}
\affiliation{%
  \institution{Department of Computer Science \\
  The University of Georgia}
  \city{Athens}
  \state{GA}
}

\author{Indrajeet Javeri}
% \authornotemark[1]
\email{indrajeet.javeri@uga.edu}
\affiliation{%
  \institution{Department of Computer Science \\
  The University of Georgia}
  \city{Athens}
  \state{GA}
}

\author{Yuan Ke}
% \authornotemark[1]
\email{yuan.ke@uga.edu}
\affiliation{%
  \institution{Department of Statistics \\
  The University of Georgia}
  \city{Athens}
  \state{GA}
}

\author{Ismailcem Arpinar}
% \authornotemark[1]
\email{budak@uga.edu}
\affiliation{%
  \institution{Department of Computer Science \\
  The University of Georgia}
  \city{Athens}
  \state{GA}
}

\author{Nicole Lazar}
% \authornotemark[1]
\email{nfl5182@psu.edu}
\affiliation{%
  \institution{Department of Statistics \\
  The Pennsylvania State University}
  \city{University Park}
  \state{PA}
}

\author{John Miller}
% \authornotemark[1]
\email{jamill@uga.edu}
\affiliation{%
  \institution{Department of Computer Science \\
  The University of Georgia}
  \city{Athens}
  \state{GA}
}

\renewcommand{\shortauthors}{Toutiaee, et al.}

\begin{abstract}
In this work, we study the pandemic course in the United States by considering national and state levels data.
We propose and compare multiple time-series prediction techniques which incorporate auxiliary variables.
One type of approach is based on spatio-temporal graph neural networks which forecast the pandemic course by utilizing a hybrid deep learning architecture and human mobility data.
Nodes in this graph represent the state-level deaths due to COVID-19, edges represent the human mobility trend and temporal edges correspond to node attributes across time.
The second approach is based on a statistical technique for COVID-19 mortality prediction in the United States that uses the SARIMA model and eXogenous variables.
We evaluate these techniques on both state and national levels COVID-19 data in the United States and claim that the SARIMA and MCP models generated forecast values by the eXogenous variables can enrich the underlying model to capture complexity in respectively national and state levels data.
We demonstrate significant enhancement in the forecasting accuracy for a COVID-19 dataset, with a maximum improvement in forecasting accuracy by \textbf{64.58\%} and \textbf{59.18\%} (on average) over the GCN-LSTM model in the national level data, and \textbf{58.79\% } and \textbf{52.40\%} (on average) over the GCN-LSTM model in the state level data.
Additionally, our proposed model outperforms a parallel study (AUG-NN) by \textbf{27.35\%} improvement of accuracy on average.

\end{abstract}



\maketitle

\section{Introduction}
The outbreak of the COVID-19 pandemic from early 2020 until today has resulted in over 189M infected individuals and in over 4M deaths worldwide ~\cite{worldhealthorganization}. 
The ability to forecast the number of infections and deaths is vital to policymakers since they can manage healthcare resources, control disease upsurges, and take preventive actions when necessary to ensure public health safety.
As the number of mortality and morbidity in the U.S. continued to rise within 2020, many states enforced the lockdown policy, practiced remote working, and imposed social distancing to slow the spread of COVID-19.
These policies also affect individual mobility. 

On the other hand, the number of cases in hospitals, ICUs, and on ventilators also increase as a result of the pandemic.
Therefore, mobility and hospitalization patterns both at the national and local levels can provide useful measures for predicting the pandemic course, especially when this data is included in the pandemic analysis \cite{kapoor2020examining}.

%\subsection{Related Work}
A great amount of research has been conducted since the beginning of the COVID-19 pandemic on forecasting the number of people affected. 
Early studies such as the one proposed in \cite{Barmparis_2020} used the SIR model for forecasting the infection rate in different countries. 
Other work \cite{fazeli2020statistical} used ARIMA to predict the daily death rate in different states in the United States. 
\cite{javeri2021improving} introduced AUG-NN model that enriches neural network models by augmentation which resulted in significant improvements in the accuracy.
They reported the forecast values on the national level data.

Time-series forecasting using Graph Neural Networks (GNN) has been introduced in various domains in the past, however, it is less studied in epidemic disease.
For example, \cite{zhao2019t} used a Temporal Graph Convolutional Network (T-GCN) for traffic prediction while \cite{matsunaga2019exploring} used Graph Neural Networks for Stock Market Prediction. 
GNN based approach for COVID-19 prediction discussed in \cite{kapoor2020examining}, generated forecast values by using spatio-temporal mobility data.
Although their work is impressive, they reported COVID-19 forecasting on a very small scale, i.e. top 20 most populated counties in the United States.

In this work, we extend the current research of pandemic modeling by proposing novel approaches for predicting daily death cases in the United States on both national and state levels.
We introduce a spatio-temporal graph convolutional network that can capture complex dynamics by including mobility patterns across different states and a statistical model that generates forecasts by eXogenous variables.
With extensive experiments among proposed methods, we demonstrate the power of eXogenous variables combined with lagged variables within the predictive models and conclude with an analysis of eXogenous variables and their potential in monitoring virus spread.

\section{Datasets}

\subsection{Aggregate Mobility Data}
The mobility data used in the study is obtained from COVID-19 U.S. Flows~\cite{kang2020multiscale}. 
This dataset consists of dynamic human mobility patterns across the United States in the form of the daily and weekly population flows at three geographic scales: census tract, county, and state.
The spatio-temporal data is obtained by analyzing, computing, and aggregating the millions of anonymous mobile phone users' visit trajectories to various places provided by ``SafeGraph''. 
To be specific, we use the daily and weekly state-to-state daily population flow~\cite{geods} starting from January 19, 2020 to January 19, 2021.
The data files consist of the unique identifiers, latitudes and longitudes for the origin and destination states, date, visitor flows (estimated number of visitors detected by SafeGraph between two geographic units), and the population flow (estimated population flows between two geographic units, inferred from visitor flows). 
 
\subsection{COVID-19 Dataset}
% The {\sc ScalaTion} COVID-19 dataset available on Github comprises data about COVID-19 cases, hospitalizations, deaths, etc. in the United States, both at the national \footnote{\url{https://github.com/scalation/data/blob/master/COVID/CLEANED_35_Updated.csv}} and state \footnote{\url{https://github.com/scalation/data/blob/master/COVID/Until\%201-14-21/USCOVID_BY_STATE.csv}} levels.
% The {\sc ScalaTion} COVID-19 dataset available on Github comprises data about COVID-19 cases, hospitalizations, deaths, etc. in the United States, both at the national \footnote{\url{https://github.com/...}} and state \footnote{\url{https://github.com/...}} levels. 

The COVID-19 dataset available on GitHub comprises data about COVID-19 cases and deaths in the United States, both at the national and state levels. 
It also contains several eXogenous variables such as "hospitalizedCurrently", "inIcuCurrently", "onVentilatorCurrently", etc.\footnote{All code and data available at \url{https://github.com/mh2t/COVID_Forecasting_Using_eX_Vars}}

The national-level data spans from January 13, 2020 to March 7, 2021, which consists of 420 days. We ignore the first 44 days due to missing values and start from February 26, 2020, when the first  COVID-19 death in the U.S. was recorded. As a result,  the national level time-series studied in this paper contains 376 days. 
We used the first 236 days as our training set and the rest 140 days as the test set.
The state-level data spans from January 19, 2020 to January 19, 2021, which consists of 367 days. Similarly, we ignore the first 70 days and start from March 29, 2020 to avoid missing values. Then, the state-level time-series studied in this paper contains 297 days for each state. We used the first 185 days as the initial training set and implement a two-week ahead rolling window forecast to assess the performance of various methods.

\section{Methodology}

\subsection{SARIMAX}
\begin{figure}[h]
 \centering
 \includegraphics[scale=0.6]{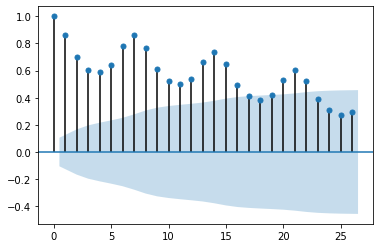}
 \caption{ACF plot of the daily death increase}
 \label{acf}
\end{figure}

Weekly seasonality of daily death increase can be observed using Auto-Correlation Function (ACF), as shown in Figure~\ref{acf}. 
Then, we propose to fit the daily death increase by a Seasonal Autoregressive Integrated Moving Average (SARIMA) \cite{arunraj2016application} model defined as below
\begin{equation}
    \label{eq:sarima}
    \varphi_{p}(B) \Phi_{P}\left(B^{s}\right) \nabla^{d} \nabla_{s}^{D} y_{t}=\theta_{q}(B) \Theta_{Q}\left(B^{s}\right) \varepsilon_{t},
\end{equation}
where $y_t$ is a variable to forecast, i.e., the logarithm of \textit{deathIncrease}, $t=1,2,\dots$, $\varphi_{p}(B)$ is a regular AR polynomial of order $p$, $\theta_{q}(B)$ is a regular MA polynomial of order $q$, $\Phi_{P}\left(B^{s}\right)$ is a seasonal AR polynomial of order $P$, and $\Theta_{Q}\left(B^{s}\right)$ is a seasonal MA polynomial of order $Q$. 
The differencing operator $\nabla^{d}$ and the seasonal differencing operator $\nabla_{s}^{D}$ eliminate the non-seasonal and seasonal non-stationarity, respectively. 

The SARIMA with eXogenous factor (SARIMAX) model is an extension of the SARIMA model in (\ref{eq:sarima}), which has the ability to include eXogenous variables, such as hospitalization and ICU occupancy rates. The SARIMAX model can be defined as:
\begin{equation}
    \label{eq:sarimax}
    \varphi_{p}(B) \Phi_{P}\left(B^{s}\right) \nabla^{d} \nabla_{s}^{D} y_{t}=\theta_{q}(B) \Theta_{Q}\left(B^{s}\right) \varepsilon_{t} + \sum^n_{i=1}\beta_i x_t^i,
\end{equation}
where $\{x_t^1,\dots,x_t^n\}$ are the $n$ eXogenous variables defined at time $t$ with coefficients $\{\beta_1,\dots,\beta_n\}$. Further, we apply a $\log$ transformation to categorical variables.

\subsection{Minimax Concave Penalty}
\label{sec:mcp}
%An alternative approach to predicting the time-series variable is to fit the regression of the past values and exogenous explanatory variables such as hospitalization.
We also consider a penalized linear regression approach to predict the daily death increase by historical data and eXogenous explanatory variables. We denote 
$
z_t=\log y_t-\log y_{(t-7)}
$
as the weekly log-return of \textit{deathIncrease} at time $t$ and $\bf{x}_t$ the eXogenous hospitalization variables at time $t$.
We linearly regress $z_t$ on  $\{z_{t-h},\dots,z_{t-(h+k-1)}\}$ and  $\{\mathbf{x}_{t-h},\dots,\mathbf{x}_{t-(h+k-1)}\}$, where $k = 14$ and $1\leq h \leq 14$ is a horizon parameter.
The \textit{horizon} refers to the number of days into the future for which forecast values are to be generated.
Since the state level hospitalization variables are highly correlated, we first implement a sure independent screening \cite{fan2008sure} procedure to reduce the dimensionality of $\bf{x}_t$. As a result, only the top 7 eXogenous variables are included in the following penalized regression model:
\begin{equation}
    \label{eq:mcp_loss}
    Q(\boldsymbol{\beta} \mid \mathbf{X}, \mathbf{z})=\frac{1}{2 N}\|\mathbf{z}-\mathbf{X} \boldsymbol{\beta}\|^{2}+\sum_{j=1}^{d} P_{\gamma} \left(\beta_j ; \lambda\right),
\end{equation}
where $\mathbf{z} \in \mathbb{R}^n$ is the vector of response variables,  $\mathbf{X} \in \mathbb{R}^{n\times d}$ is the design matrix of all predictors and $\boldsymbol\beta = (\beta_1,...,\beta_{d})^{\mathrm{T}}$ is a vector of unknown regression coefficients. Besides,  $P_{\gamma}(\cdot;\lambda)$ is the Minimax Concave Penalty (MCP) \cite{zhang2010nearly} which satisfies
\begin{equation}
    \label{eq:mcp}
P_{\gamma}(\beta ; \lambda)=\left\{\begin{array}{ll}
\lambda|x|-\frac{\beta^{2}}{2 \gamma}, & \text { if }|\beta| \leq \gamma \lambda, \\
\frac{1}{2} \gamma \lambda^{2}, & \text { if }|\beta|>\gamma \lambda.
\end{array}\right.
\end{equation}

\iffalse
Since $(n+1)\times k$ explanatory variables in the regression model tend to overfit, the number of predictors has been reduced from $(n+1)\times k$ to $m=7$. 
The $m=7$ explanatory variables are the top $m$ ones, which have larger marginal Pearson's correlation coefficients with the response variable.

After excluding a certain number of explanatory variables by Pearson's correlation coefficient, we used the Minimax Concave Penalty (MCP) \cite{zhang2010nearly} to further select variables by data. 
MCP selects variables by minimizing the penalized squared loss via:
\begin{equation}
    \label{eq:mcp_loss}
    Q(\boldsymbol{\beta} \mid \mathbf{X}, \mathbf{z})=\frac{1}{2 N}\|\mathbf{z}-\mathbf{X} \boldsymbol{\beta}\|^{2}+\sum_{j=1}^{m} P_{\gamma} \left(\beta_j ; \lambda\right),
\end{equation}
with the minimax concave penalty:
\begin{equation}
    \label{eq:mcp}
P_{\gamma}(x ; \lambda)=\left\{\begin{array}{ll}
\lambda|x|-\frac{x^{2}}{2 \gamma}, & \text { if }|x| \leq \gamma \lambda \\
\frac{1}{2} \gamma \lambda^{2}, & \text { if }|x|>\gamma \lambda
\end{array}\right.
\end{equation}
indexed by $\gamma>1$, where $\boldsymbol\beta = \{\beta_1,...,\beta_{m}\}$ is the coefficient vector in the linear model and $N$ is the sample size. 
\fi

\subsection{Vector Auto-Regressive Model}
 Vector Auto-Regressive model \cite{zivot2006vector} is another popular approach to model and predicts multivariate time-series. 
 For a $p$ dimensional response vector of interest, say: 
\begin{equation}
\mathbf{y}_t = (y_{1,t},y_{2,t},\ldots, y_{n,t})^{\mathrm{T}},
\end{equation}
a vector auto-regressive model of order $q$, i.e. VAR(q), is defined as
\begin{equation}
\mathbf{y}_t = \delta +\Phi^{(0)}\mathbf{y}_{t-q} + \Phi^{(1)}\mathbf{y}_{t-q+1} + \ldots + \Phi^{(q-1)}\mathbf{y}_{t-1} + \epsilon_t
\end{equation}
where  $\mathbf{\delta} \in \mathbb{R}^n$ is an intercept vector, $\Phi^{(s)} \in \mathbb{R}^{n\times n} (s = 0, \ldots, p - 1)$ are regression coefficient matrices, and $\epsilon_t \in \mathbb{R}^n$ is an error vector.

\iffalse
If multiple variables are involved in forecasting, Multivariate Time-Series Analysis (MTSA) can be used.
MTSA extends univariate time-series by analyzing multiple variables. 
It works on multiple interrelated time-series,
\begin{equation}
\mathbf{y}_t = \{y_{t0},y_{t1},\ldots, y_{t,n-1}\}
\end{equation}
\noindent with one time-series for each of the $n$ variables.
The forecasted value for the $j^{th}$ variable at time t, $y_{tj}$, can depend on the previous (or lagged) values of all the variables.
This notion is captured in Vector Auto-Regressive models \cite{zivot2006vector}.
A vector auto-regressive model of order $p$ with $n$ variables, VAR$(p,n)$, will utilize the most recent $p$ values for each variable to produce a forecast.
The equation for the response vector $\mathbf{y}_t \in \mathbb{R}^n$ can be written in matrix-vector form as follows:
\begin{equation}
\mathbf{y}_t = \delta +\Phi^{(0)}\mathbf{y}_{t-p} + \Phi^{(1)}\mathbf{y}_{t-p+1} + \ldots + \Phi^{(p-1)}\mathbf{y}_{t-1} + \epsilon_t
\end{equation}
\noindent where constant vector $\bm{\delta} \in \mathbb{R}^n$, parameter matrices $\Phi^{(l)} \in \mathbb{R}^{n\times n} (l = 0, \ldots, p - 1)$, and error vector $\epsilon_t \in \mathbb{R}^n$.
\fi

Similar to the data pre-processing procedure described in Section~\ref{sec:mcp}, a weekly log-return has been applied to both the response vector and eXogenous hospitalization variables to remove the seasonality.

\subsection{Random Walk}
By definition, a candidate series follows a random walk if the first differences are random (non-stationary).
Random Walk (RW) is a common technique in graphical models \cite{aldous2014reversible}, and it is widely used in webpage ranking, image segmentation, and time-series analysis.
The most practical usage of RW is in financial markets, where it states that the historical trend of a market cannot be used to forecast its future trend.
Many studies showed that the RW method is applicable to most time-series data, especially when the samples have the same distribution and are independent of each other.
A Gaussian Random Walk for variable $y_t$ can be written by:
\begin{equation}
y_t = y_{t-1} +\epsilon_t 
\end{equation}
where $\epsilon_t$ follows Gaussian distribution.
% \noindent A stochastic process $\{\epsilon_t | t \in \{0,\ldots, m-1\}\}$ is considered to be Gaussian if:
% \begin{align}
% \mathbb{E}[\epsilon_t] & = 0  &\text{zero mean}\\
% \mathbb{E}[\epsilon_t^2] & = \sigma^2  &\text{constant variance}\\
% \mathbb{E}[\epsilon_t\epsilon_{t-1}] & = 0  &\text{uncorrelated}
% \end{align}
% \noindent Several distributions can be used to generate white noise with the most common being the Gaussian (Normal) distribution, $N(0,\sigma^2I)$.

\subsection{GCN-LSTM}
\textbf{GCN:} The main idea behind Graph Convolutional Network (GCN) models is that a given input signal (node) can be enriched via information propagation from its neighbors to improve a future prediction task.
The neighbors are often defined by constructing a network of inputs where nodes and connections represent features and relations between them.
This study~\cite{gilmer2017neural} formulated an oracle definition upon the message-passing framework that was inspired by many previously proposed methods.
In such settings, the spatial dependency is obtained by the 2-layer GCN architecture through the following formula:
% \begin{equation}
% \bm{m}_i^{(l+1)} = \sum_{j\in \mathcal{N}(i)} \mathcal{F}^{(l)} \left(\bm{h}_i^{(l)},\bm{h}_j^{(l)}\right),\quad \bm{h}_i^{(l+1)}=\mathcal{G}^{(l)} \left(\bm{h}_i^{(l)},\bm{m}_i^{(l+1)}\right)
% \end{equation}
% where $\mathcal{F}^{(l)}$ denotes message function, $\mathcal{G}^{(l)}$ denotes node update function, $\bm{m}_i^{(l)}$ denotes the messages propagated between nodes, and $\bm{h}_i^{(l)}$ denotes the node representations.
% This process is implemented in two steps:
% first, the information follow along the neighbors; and second, the information are aggregated to determine the updated hidden representations.
\begin{equation}
f(X,A) = \sigma (\hat{A}\:Relu(\hat{A}\,X\,W_0)\,W_1)    
\end{equation}
where $X$ denotes the feature matrix, $A$ denotes the adjacency matrix, $W_0$ and $W_1$ respectively denote the weight matrix in the first and second layer.  

\noindent\textbf{Mobility Network Graph:}
In pandemic modeling, we usually analyze the global model through multiple time-series data obtained at the local level, which denotes the spreading dynamics in each region.
The forecasting task is usually defined as a regression model that takes in a time-series of $t-k,\ldots,t-1,t$ and emits a single value $t+1$.
In a general case, the output can be a series of future time points $t+1,t+2,\ldots$ as generated forecast values.
However, the regression model requires an adjustment for modeling human mobility across regions.
Mobility data forms a spatial-graph, where a region $i$ is denoted by nodes, and every node can be connected to other nodes $j,k,z,\ldots$, and weighted edges represent the strength of relations between the nodes.

\noindent\textbf{Binary Graph}: We constructed a binary adjacency matrix to feed into the trainer.
This adjacency matrix has been created by the following steps:
(1) the average of mobility data along the time point has been obtained, and
(2) if the number of movements from origins to a destination is among the top $20\%$, the corresponding cell was assigned 1, and 0 otherwise.
(3) Finally, the matrix was corrected to be a full-rank matrix by an orthonormal set obtained via the Gram—Schmidt process.
We used \verb|matlib| package in R for this purpose.

\noindent \textbf{GCN-LSTM:} Disease epidemic forecasting is a quintessential example of spatio-temporal problems for which we present a deep neural network framework that captures the number of deaths using spatio-temporal data. 
The task is challenging due to two main inter-twined factors: 
(1) the complex spatial dependency between time-series of each state, and 
(2) non-linear temporal dynamics with changing non-pharmaceutical interventions (NPI) such as mobility trends.
%%%%%%%%%%%%%%%%%%%%%%%%%%%%%%%%%%%%%%%%%%%%%%%%%%%%%
%GCN_LSTM
% \begin{figure}[h]
%   \centering
%   \includegraphics[width=\linewidth]{plot/gcn.png}
%   \caption{An overview of GCN\_LSTM architecture.}\label{fig:model}
% \end{figure}
%%%%%%%%%%%%%%%%%%%%%%%%%%%%%%%%%%%%%%%%%%%%%%%%%%%%%

We attempt to populate a temporal knowledge graph using mobility patterns, since people moving around the regions with similar epidemic patterns may contribute more to the forecasting process.
The people traveling serve as the ground truth for training a GCN for identifying the underlying graph between sub-regions.
Next, the constructed graph embedding from the GCN model is used to feed into an LSTM model to forecast the pandemic in the future.
Notice that the graph embedding provides knowledge about the forecasting system, and the LSTM provides a direction for how to leverage the GCN output in the COVID-19 forecasting task.

\subsection{Rolling Validation for Multiple Horizons}\label{sec:rolling}
Classical multi-folds cross-validation approaches are not directly applicable to time-series data due to the existence of serial dependence. 
Instead, we follow a rolling-validation scheme in this study. To be specific, we reserve the first 60\% of days in the time-series as the training set and use the rest 40\% of days to make a rolling window forecast. 
For example, a forecast with horizon parameter $1\leq h \leq 14$ is obtained using the model trained in the training, an input from a rolling window contains the information from $t-13$ to time $t$, and an out-of-sample forecast for time $t+h$. 
Then, we move the rolling window forward by two weeks and repeat the above process. The two weeks ahead forecast policy is suggested by CDC.
The forecast accuracy is measured by the symmetric mean absolute percentage error (sMAPE) which the smaller the better.

\iffalse
Cross-validation in time-series data is not applicable with the classic form, because there are strong dependencies between instances and their neighbors in time.
The principle of splitting the data into train and test sets is also applied in a simple form of rolling-validation.
For example, we separated 60\% of the samples as the training set in our analysis and the rest as the test set.
Horizon $h=1$ forecasting is obtained based on the forecast value generated by training on the training set.
The error is the difference between the true value in the test set and the forecast value, and it is the basis of the symmetric mean absolute percentage error (sMAPE).

For the purpose of simplicity and efficiency in models, the process of rolling forward to forecast the next value in the test set often involves retraining the model by including the first value in the test set in the training set.
The first value is then removed from the training set, so the size of the training set remains the same.
We trained our architecture only once on the training set to generate the out-of-sample forecast values.
\fi

\section{Experiments}

\subsection{Hyperparameters and Architectures}
\noindent\textbf{GCN-LSTM:}
The GCN-LSTM architecture used in this experiment consists of two GCN layers followed by one LSTM layer. The model is built using \verb|StellarGraph| library ~\cite{StellarGraph}. The sizes of each of these layers vary for each horizon. The size of GCN layers falls between 10 and 32 while the size of LSTM layers falls between 150 and 300. Information from 10 previous lags is used for forecasting future instances.
 
\noindent\textbf{SARIMAX:}
%The exogenous variables used in SARIMAX are the number of daily cases in hospitals and ICUs. 
The eXogenous variables used in SARIMAX at both the national and state levels are "hospitalizedCurrently" and "inIcuCurrently". The specific model we fit is a SARIMAX$(4,1,4)\times(3,1,1,7)$ model with a constant trend. 

\noindent\textbf{MCP:}
We fit an MCP model and consider the serial dependence among daily new cases in hospitals.
Each day's new hospitalized count is dependent on the previous 14 days of observations.
The regularized parameters in MCP are selected by a multi-fold cross-validation.
Since the number of predictors is large, a Sure Independence Screening procedure is used to remove the uninformative predictors. 
As a result,  7 predictors are included in the MCP model.

\noindent\textbf{VAR:}
Similar to SARIMAX model, VAR model contains multiple predictors including ``inIcuCurrently'', ``hospitalizedCurrently'', ``hospitalizedCumulative'' and ``onVentilatorCurrently'' in addition to ``deathIncrease'' variable. The hyperparameters are selected by the Bayesian Information Criterion (BIC).

\subsection{Baseline}
%\noindent\textbf{RW:}
We compare the performance of all the statistical models and the GCN-LSTM model with a baseline Random Walk (RW) model. 
For each model, we make a 14-day ahead rolling window forecast as introduced in Section \ref{sec:rolling}.
The forecast performance is measured by the sMAPE score at each horizon $1\leq h \leq 14$. 

The national-level forecast results are reported in Table \ref{tab:us}. 
The SARIMAX model performs the best on this dataset with the lowest sMAPE score. RW has slightly better results than other models at horizons $h=7$ and 14.

Table \ref{tab:bystate} shows the forecast results on the state-level data. 
For each state in the U.S., we make a 14-day ahead rolling window forecast with horizons $1\leq h \leq 14$. Then, we aggregate the state-level forecast into a forecast for the national level daily death increase.
The sMAPE scores on this aggregated forecast are then used to compare all models. 
MCP model performs the best on this experiment while RW has the worst performance in most scenarios. 
%RW has slightly better results on horizon 7 and 14 on this dataset as well.

Table \ref{tab:state-all-models} reports the sMAPE scores for different models on a set of representative states. 
We notice that the scores for each model are much higher than the aggregated results reported in Table \ref{tab:bystate}. 
Also, we find that the GCN-LSTM model performs the worst when the forecasts for these representative states are considered.

\subsection{Mortality Prediction Performance}
In Tables~\ref{tab:us} and \ref{tab:bystate}, we compare the forecasting performance of SARIMAX with a range of models.
We report the sMAPE score for the predicted ``deathIncrease''.
Table~\ref{tab:us} shows the performance of models over the national-level data, and Table~\ref{tab:bystate} shows the performance over aggregated state-level data for the United States mortality.
All performance values were reported for 14 horizons.
According to Table~\ref{tab:us}, the SARIMAX model achieves the best sMAPE score at each horizon and outperforms other models on the national level forecast.
The best results were achieved by evaluating several combinations, i.e. hospitalizedIncrease, hospitalizedCurrently, etc.

On the other hand, the MCP model performs best on the state level forecast which is evidenced by the results in Table~\ref{tab:bystate}.
This trend is due to the fact that we included more lags ($k=14$) in MCP compared to SARIMAX, and MCP tends to manage overfitting better than SARIMAX since it uses a regularized term.
Although SARIMAX appears to be more accurate than MCP in forecasting national-level data, it is inferior to MCP in forecasting state-level data, owing to the fact that the global combination might not be the best option for all the states.

Further, we discovered a general pattern that the forecast results can be improved by including eXogenous variables. 
That may explain the success of SARIMAX and MCP in our study.
Interestingly, introducing additional variables resulted in worse performance for the SARIMAX model.
Table~\ref{tab:state-all-models} compares different models for selected states.

% ~~~~~~~~~~~~~~~~~~~~~~~~~~~~~~~~~~~~~~~~~~~~~~~~~~~~~~~~~~~~~~~~~~~~~~~~~~~~~~~~~~~
%% CLEANED_35_Updated.csv
% ~~~~~~~~~~~~~~~~~~~~~~~~~~~~~~~~~~~~~~~~~~~~~~~~~~~~~~~~~~~~~~~~~~~~~~~~~~~~~~~~~~~
\begin{small}
\begin{table}[h]
\begin{center}
\caption{Multi-Horizon ($h$) Rolling Forecasts for the United States (National Level Data): Competitive Models (sMAPE).}
\label{tab:us}
\begin{tabular}{|c|c|c|c|c|c|c|c|c|} \hline
Horizon  & RW & GCN & \textbf{SARIMAX} & SARIMA & MCP & VAR \\ \hline\hline
$h = 1$  & 29.20 & 28.87 & 13.25 & 16.02 & 15.61 & 15.08  \\ \hline
$h = 2$  & 48.40 & 33.81 & 15.10 & 19.36 & 17.13 & 16.09  \\ \hline
$h = 3$  & 53.50 & 35.97 & 15.47 & 19.13 & 17.85 & 16.55  \\ \hline
$h = 4$  & 54.20 & 32.26 & 15.21 & 19.26 & 18.02 & 16.53  \\ \hline
$h = 5$  & 50.10 & 29.83 & 14.98 & 20.40 & 18.02 & 16.62  \\ \hline
$h = 6$  & 32.20 & 36.76 & 14.18 & 20.49 & 17.87 & 16.50  \\ \hline
$h = 7$  & 18.20 & 34.04 & 14.03 & 20.45 & 17.86 & 16.75  \\ \hline
$h = 8$  & 31.30 & 34.29 & 13.52 & 22.70 & 23.03 & 18.61  \\ \hline
$h = 9$  & 47.40 & 48.39 & 14.06 & 24.92 & 22.94 & 19.58  \\ \hline
$h = 10$ & 52.10 & 54.37 & 14.22 & 25.77 & 23.24 & 19.84  \\ \hline
$h = 11$ & 53.60 & 32.07 & 14.25 & 25.75 & 23.39 & 19.86  \\ \hline
$h = 12$ & 49.90 & 28.39 & 14.52 & 26.27 & 23.16 & 20.10  \\ \hline
$h = 13$ & 34.20 & 35.63 & 15.05 & 27.00 & 22.91 & 19.79  \\ \hline
$h = 14$ & 24.30 & 32.84 & 15.22 & 26.94 & 22.97 & 20.17  \\ \hline
\hline
Average  & 41.33 & 35.54 & \textbf{14.50} & 22.46 & 20.29 & 18.01 \\ \hline
\end{tabular}
\end{center}
\end{table}
\end{small}
% ~~~~~~~~~~~~~~~~~~~~~~~~~~~~~~~~~~~~~~~~~~~~~~~~~~~~~~~~~~~~~~~~~~~~~~~~~~~~~~~~~~~
%% USCOVID_BY_STATE.csv
% ~~~~~~~~~~~~~~~~~~~~~~~~~~~~~~~~~~~~~~~~~~~~~~~~~~~~~~~~~~~~~~~~~~~~~~~~~~~~~~~~~~~
\begin{small}
\begin{table}[h]
\begin{center}
\caption{Multi-Horizon ($h$) Rolling Forecasts for the United States (State Level Data): Competitive Models (sMAPE).}
\label{tab:bystate}
\begin{tabular}{|c|c|c|c|c|c|c|c|c|} \hline
Horizon   & RW    & GCN   & SARIMAX & SARIMA & \textbf{MCP}     & VAR \\ \hline\hline
$h = 1$   & 29.66 & 24.99 & 18.22 & 15.41 & 16.28 & 15.40 \\ \hline
$h = 2$   & 50.17 & 34.42 & 18.01 &	16.36 & 16.09 & 15.42 \\ \hline
$h = 3$   & 53.98 & 34.75 & 18.42 &	17.31 & 16.17 & 16.26 \\ \hline
$h = 4$   & 54.30 & 34.42 & 17.53 &	17.84 & 16.38 & 16.32 \\ \hline
$h = 5$   & 51.71 & 30.41 & 17.58 & 17.95 & 16.61 & 15.90 \\ \hline
$h = 6$   & 33.89 & 30.81 & 18.73 & 17.76 & 16.51 & 16.17 \\ \hline
$h = 7$   & 17.44 & 35.74 & 18.04 & 17.84 & 16.63 & 16.65 \\ \hline
$h = 8$   & 31.54 & 34.62 & 18.64 & 21.21 & 18.17 & 19.65 \\ \hline
$h = 9$   & 47.21 & 38.88 & 18.64 & 21.89 & 17.82 & 20.67 \\ \hline
$h = 10$  & 51.13 & 44.04 & 18.52 & 22.98 & 19.86 & 21.07 \\ \hline
$h = 11$  & 52.70 & 44.15 & 19.02 & 23.24 & 19.40 & 22.36 \\ \hline
$h = 12$  & 49.94 & 43.34 & 29.08 & 23.62 & 17.98 & 20.36 \\ \hline
$h = 13$  & 34.91 & 38.21 & 26.88 & 23.63 & 18.19 & 20.31 \\ \hline
$h = 14$  & 23.30 & 44.46 & 23.10 & 23.33 & 18.32 & 20.13 \\ \hline
\hline
Average   & 41.56 & 36.66 & 20.03 & 20.81 & \textbf{17.55} & 18.33 \\ \hline

\end{tabular}
\end{center} 
\end{table}
\end{small}

\section{Conclusion and Future Work}

In this paper, we evaluated and compared several statistical and machine learning models to forecast the pandemic course in the United States, using national and state levels data.
We studied the effectiveness of the mobility data in the COVID-19 prediction problem regarding the accuracy and discussed the benefits of including eXogenous variables.
The advantage of SARIMAX and MCP, over other competitive methods such as GCN-LSTM, is that they can incorporate eXogenous variables such as hospitalization, ICU occupancy rate, and the count of patients who require a ventilator mask into a multivariate time-series forecast. 
The empirical advantages of including such eXogenous variables were justified by our experiments. 

Our work adds to the growing body of epidemic disease modeling with a novel approach to combine the multivariate time-series analysis with human mobility data and eXogenous variables.
We expect the findings in this paper can motivate future studies for selecting and incorporating important eXogenous variables into time-series modeling to enhance the prediction.
\newpage
\bibliographystyle{ACM-Reference-Format}
\bibliography{main}

\begin{appendix}
\section{Appendix}
\begin{small}
\begin{table*}[h]
\begin{center}
\caption{Multi-Horizon ($h$) Rolling Forecasts for Selected States: Random Walk, SARIMA, GCN-LSTM, SARIMAX, VAR and MCP (sMAPE). }
\label{tab:state-all-models}
\begin{tabular}{|c|c|c|c|c|c|c|c|c|c|c|c|c|c|c|c|c|c|c|c|c|c|} \hline
  State & \multicolumn{6}{c|}{CA} & \multicolumn{6}{c|}{GA} & \multicolumn{6}{c|}{IL} \\ \hline\hline
Horizons  & RW    & SR    & GCN   &  \textbf{SRX}  & VAR   & MCP   &  RW   & \textbf{SR}    & GCN   &  SRX  & VAR   & MCP   &  RW   & SR    & GCN   &  \textbf{SRX}  & VAR   & MCP\\ \hline\hline
$h = 1$   & 54.10 & 38.60 & 55.63 &  35.63&  42.02&  35.10& 68.80 & 54.40 & 58.74 &  59.54&  51.87&  53.13& 41.60 & 27.40 & 61.76 &  30.49&  32.72& 32.99  \\ \hline
$h = 2$   & 68.50 & 39.40 & 57.86 &  38.02& 41.06 &  34.42& 82.00 & 51.50 & 62.68 &  57.52&  51.93&  55.08& 53.00 & 29.00 & 64.67 &  29.25&  31.63& 32.73  \\ \hline
$h = 3$   & 76.20 & 40.80 & 64.93 &  36.44&  41.11&  33.93& 87.00 & 52.80 & 68.37 &  59.93&  51.09&  52.49& 54.90 & 30.10 & 68.11 &  30.05&  31.67&  32.43  \\ \hline
$h = 4$   & 75.90 & 41.70 & 68.76 & 37.50 &41.06  &  33.70& 87.60 & 52.60 & 70.51 &  56.92&  51.06&  53.57& 55.10 & 31.00 & 70.67 &  29.44&  31.38& 32.97   \\ \hline
$h = 5$   & 66.10 & 41.70 & 68.31 & 35.31 & 41.32 & 34.34 & 83.30 & 52.20 & 70.14 &  56.42&  51.18&  53.45& 52.10 & 32.30 & 71.32 &  29.17&  31.07& 32.84   \\ \hline
$h = 6$   & 55.60 & 40.60 & 65.53 &  38.44& 41.40 &  34.98& 69.10 & 51.60 & 69.05 &  56.36&  51.13&  53.62& 43.20 & 31.70 & 70.93 &  30.76&  31.57& 31.94   \\ \hline
$h = 7$   & 41.40 & 40.90 & 60.42 &  40.54& 41.48 &  34.73& 48.80 & 52.50 & 62.67 &  54.57&  51.54&  53.77& 31.00 & 31.40 & 70.04 &  30.60&  31.47& 31.78   \\ \hline
$h = 8$   & 59.60 & 47.30 & 55.34 &  37.00& 42.62 &  41.94& 69.70 & 56.30 & 59.24 &  61.46&  61.13&  63.82& 40.40 & 35.20 & 70.57 &  30.52&  34.45& 36.89   \\ \hline
$h = 9$   & 68.80 & 47.90 & 58.54 &  37.55& 43.28 &  42.44& 79.70 & 56.80 & 63.7  &  61.21&  60.27&  60.75& 51.10 & 37.00 & 73.42 &  31.34&  33.21& 34.82   \\ \hline
$h = 10$  & 74.70 & 49.90 & 66.69 &  42.21&  43.44&  42.46& 82.40 & 57.60 & 67.85 &  61.98&  60.85&  60.81& 54.50 & 38.10 & 75.41 &  31.89&  33.47& 39.67   \\ \hline
$h = 11$  & 74.50 & 50.60 & 68.88 &  36.79&  43.40&  43.02& 83.70 & 57.30 & 72.01 &  58.91&  60.37&  62.04& 56.20 & 38.50 & 76.99 &  29.48&  34.41& 36.63   \\ \hline
$h = 12$  & 67.80 & 51.40 & 67.8  &  39.08&  43.75&  43.10& 79.50 & 57.00 & 73    &  59.55&  60.56&  59.43& 52.70 & 39.90 & 78.53 &  30.19&  33.87& 35.09   \\ \hline
$h = 13$  & 57.40 & 49.70 & 64.91 &  37.79&  43.93&  42.87& 66.10 & 55.80 & 71.01 &  59.46&  60.45&  60.58& 42.40 & 41.10 & 78.42 &  31.59&  34.64& 35.03   \\ \hline
$h = 14$  & 48.60 & 50.40 & 59.42 &  34.93&  44.37&  43.80& 55.50 & 56.20 & 62.24 &  58.79&  60.74&  61.46& 36.20 & 39.80 & 77.51 &  32.84&  35.68& 34.83   \\ \hline\hline

Average &  63.51 &	45.06 &	63.07 &	\textbf{37.66} &	42.45&	38.63 &	74.51&	\textbf{54.61}&	66.52 &	58.76&	56.01&	57.43&	47.46&	34.46&	72.03&	\textbf{30.54}&	32.95&	34.33 \\ \hline\hline

State & \multicolumn{6}{c|}{TX} & \multicolumn{6}{c|}{NY} & \multicolumn{6}{c|}{PA} \\ \hline\hline

Horizons  & RW    & SR    & GCN   &  SRX  & \textbf{VAR}   & MCP   &   RW  & SR    & GCN   &  \textbf{SRX}  & VAR   & MCP   &  RW   & SR    & GCN   &  \textbf{SRX}  & VAR   & MCP          \\ \hline\hline
$h = 1$   & 58.80 & 29.00 & 36.01 &  27.88&  28.19&  29.53& 26.50 & 25.70 & 35.04 &  20.76&  24.06&  24.62& 57.20 & 46.40 & 70.86 &  45.86& 46.30 & 42.96      \\ \hline
$h = 2$   & 76.40 & 32.60 & 57.29 &  25.92&  27.50&  29.66& 25.20 & 28.20 & 36.56 &  20.47&  24.46&  25.79& 69.70 & 47.20 & 74.13 &  46.85&  44.87& 42.62      \\ \hline
$h = 3$   & 84.30 & 33.30 & 70.94 &  32.02&  27.76&  29.43& 24.50 & 32.80 & 37.84 &  20.25&  24.56&  25.45& 78.50 & 47.00 & 77.8  & 44.13 &  45.24&42.97\\ \hline
$h = 4$   & 84.30 & 33.00 & 74.1  &  26.16&  27.58&  28.57& 27.30 & 37.50 & 39.69 &  21.92&  24.28&  24.93& 79.00 & 46.60 & 80.19 &  44.50&  46.56& 41.88      \\ \hline
$h = 5$   & 75.60 & 32.60 & 74.4  &  34.15&  27.86&  28.39& 26.00 & 40.40 & 41.71 &  22.99&  24.65&  25.08& 75.80 & 46.90 & 80.9  &  45.30&  46.23& 41.69      \\ \hline
$h = 6$   & 60.70 & 32.60 & 62.65 &  36.06&  27.59&  28.42& 30.00 & 28.70 & 42.72 &  23.61& 24.41 &  25.80& 61.60 & 46.30 & 80.57 &  44.05&  46.26&  42.61     \\ \hline
$h = 7$   & 27.20 & 32.70 & 47.52 &  28.22&  27.36&  28.43& 27.90 & 31.90 & 44.08 &  23.31&  25.22& 27.32 & 44.30 & 46.80 & 79.94 &  43.93&  45.99&  42.14      \\ \hline
$h = 8$   & 57.60 & 39.40 & 38.89 &  33.59&  33.47&  31.64& 31.10 & 40.10 & 45.72 &  33.50&  31.00&  30.09& 55.10 & 51.30 & 80.92 &  46.75&  51.03&  51.61      \\ \hline
$h = 9$   & 74.20 & 41.40 & 58.74 &  30.72&  32.79&  32.58& 33.40 & 44.20 & 47.83 &  31.02&  31.54&  29.33& 69.30 & 52.70 & 82.98 &  46.59&  51.03&  55.01      \\ \hline
$h = 10$  & 79.90 & 42.20 & 72.86 &  30.80& 32.76 &  32.26& 34.20 & 49.40 & 47.65 &  42.86& 30.72 &  27.41& 72.50 & 52.90 & 84.92 &  44.69&  49.29&  57.48      \\ \hline
$h = 11$  & 79.60 & 41.90 & 76.34 &  30.62&  32.68&  32.37& 37.60 & 55.00 & 49.62 &  29.57&  30.49&  30.28& 76.00 & 53.60 & 87.59 &  46.79&  48.18&  55.89      \\ \hline
$h = 12$  & 72.00 & 41.60 & 76.26 &  41.85&  32.58&  32.11& 37.60 & 57.80 & 52.95 &  28.17&  30.20&  32.02& 71.40 & 55.80 & 87.83 &  46.22&  49.90&  54.43      \\ \hline
$h = 13$  & 60.30 & 41.50 & 64.47 &  35.27& 32.52 &  32.17& 40.30 & 46.30 & 53.56 &  28.31&  30.89&  32.96& 61.80 & 54.60 & 87.33 &  44.91&  48.56&  52.90      \\ \hline
$h = 14$  & 33.80 & 41.40 & 50.34 &  27.48&  32.13&  32.25& 42.00 & 51.30 & 55.03 &  32.02&  31.59&  32.92& 48.90 & 55.10 & 87.83 &  45.80&  46.71  &   54.30     \\ \hline\hline

Average &  66.05 &	36.80&	61.49&	31.48&	\textbf{30.20}&	30.56&	31.69&	40.66&	45.00&	\textbf{27.05}&	27.72&	28.14&	65.79&	50.23&	81.70&	\textbf{45.46}&	47.58&	48.46 \\ \hline
\end{tabular}
\end{center}
\end{table*}
\end{small}
\end{appendix}

\end{document}